\documentclass[conference]{IEEEtran}
\IEEEoverridecommandlockouts
\usepackage{cite}
\usepackage{url}
\usepackage{colortbl} 
\usepackage{longtable}
\usepackage{amsmath,amssymb,amsfonts}
\usepackage{algorithmic}
\usepackage{multirow}
\usepackage{graphicx}
\usepackage{textcomp}
\usepackage{xcolor}
\def\BibTeX{{\rm B\kern-.05em{\sc i\kern-.025em b}\kern-.08em
    T\kern-.1667em\lower.7ex\hbox{E}\kern-.125emX}}
\begin{document}

\title{PIPES: A Meta-dataset of Machine Learning Pipelines \\ }
\author{
Cynthia Moreira Maia\textsuperscript{\textdagger} \quad
Lucas B. V. de Amorim\textsuperscript{\textdagger\textasteriskcentered} \quad
George D. C. Cavalcanti\textsuperscript{\textdagger} \quad
Rafael M. O. Cruz\textsuperscript{\textdaggerdbl} \\[0.5em]
\textsuperscript{\textdagger}Centro de Informática, Universidade Federal de Pernambuco, Recife, Brazil\\
\textsuperscript{\textasteriskcentered}Instituto de Computação, Universidade Federal de Alagoas, Maceió, Brazil\\
\textsuperscript{\textdaggerdbl}École de Technologie Supérieure, Université du Québec, Montréal, Canada\\[0.5em]
\{cmm3, lbva, gdcc\}@cin.ufpe.br, lucas@ic.ufal.br, rafael.menelau-cruz@etsmtl.ca
}

%\and
%\and

\maketitle

\begin{abstract}

Solutions to the Algorithm Selection Problem (ASP) in machine learning face the challenge of high computational costs associated with evaluating various algorithms' performances on a given dataset. To mitigate this cost, the meta-learning field can leverage previously executed experiments shared in online repositories such as OpenML. OpenML provides an extensive collection of machine learning experiments. However, an analysis of OpenML's records reveals limitations. It lacks diversity in pipelines, specifically when exploring data preprocessing steps/blocks, such as scaling or imputation, resulting in limited representation. Its experiments are often focused on a few popular techniques within each pipeline block, leading to an imbalanced sample. To overcome the observed limitations of OpenML, we propose PIPES, a collection of experiments involving multiple pipelines designed to represent all combinations of the selected sets of techniques, aiming at diversity and completeness. PIPES stores the results of experiments performed applying 9,408 pipelines to 300 datasets. It includes detailed information on the pipeline blocks, training and testing times, predictions, performances, and the eventual error messages. This comprehensive collection of results allows researchers to perform analyses across diverse and representative pipelines and datasets. PIPES also offers potential for expansion, as additional data and experiments can be incorporated to support the meta-learning community further. The data, code, supplementary material, and all experiments can be found at https://github.com/cynthiamaia/PIPES.git.

%One meta-learning (MtL) challenge is the cost of evaluating models' performance on datasets. Leveraging historical data from repositories that share machine learning experiments can mitigate this challenge. However, when analyzing one of the largest public repositories, OpenML, we observed limitations that challenge experiments' reproduction, particularly concerning pre-processing techniques. Therefore, we propose MetaD-MLP, a robust meta-dataset that can benefit the meta-learning area with a collection of experiments covering pipeline blocks, such as scaling, feature pre-processing, imputation, and encoding, covering the observed limitations of OpenML. The MetaD-MLP experiments were analyzed in 200 datasets with 9,408 combinations, with complete information about failed blocks, training time and test, and all combination predictions. Furthermore, a collection of extracted meta-features is made available, covering six types of meta-features. The experiments are available and coded on the project website
%\footnote{\url{https://drive.google.com/drive/folders/17ZOR4tbd5ML_-bCzvNuKAbenq8yT6UmZ?usp=drive_link}}.
\end{abstract}


\section{Introduction}

%The reproducibility of experiments carried out in machine learning, with the disclosure of all resources used and a clear understanding of the decisions made in the experiments, is a challenge. The lack of reproducibility limits advances in the area due to the difficulty of reusing, comparing experiments, and ends up requiring time and effort that could be avoided with standardized, easy-to-use, easy-to-compare, and reliable tools, collections of experiments \cite{Vanschoren, Brazdil}. Current machine learning support platforms still face limitations in offering simple solutions that facilitate the reproducibility of experiments. Despite advances, many tools are not ready for immediate use \cite{Gundersen}. And studies reinforce the non-reproducibility \cite{Olorisade}, \cite{Hutson}.

The Algorithm Selection Problem (ASP) was formulated by Rice in 1976 and refers to the challenge of choosing the most appropriate algorithm to solve a specific problem, considering several algorithms available \cite{rice1976algorithm}. ASP is one of the research focuses in the area of Meta-Learning (MtL) \cite{smith2009cross, song2012automatic, de2008ranking,pimentel2019new,ferrari2015clustering,khan2020literature}, a field that allows to learn from previous machine learning experiences and transfer the acquired knowledge to new \cite{hutter2019automated} tasks. MtL achieves this by mapping the characteristics (meta-features) of datasets with information describing the performance of algorithms. However, a limitation in this field is the high computational cost associated with evaluating algorithms' performances on datasets and extracting meta-features \cite{khan2020literature}. A promising approach to alleviate these costs is to use machine learning experiments available in public online repositories, where hundreds of experiments are released to the community, with a large set of machine learning techniques, which can be used to facilitate reuse and provide faster advancement in the area \cite{brazdil2022metalearning}. The effectiveness of these repositories depends heavily on community contributions, mainly through complete and efficient experiment registries. These shared registries help reduce the computational burden associated with individual experimentation.

%MtL allows you to learn from previous machine learning experiences and transfer the acquired knowledge to new \cite{hutter2019automated} tasks, 

%mapping the characteristics (meta-features) of datasets with information describing the performance of algorithms. However, a limitation in this field is the high computational cost associated with evaluating algorithms' performance on datasets and extracting meta-features \cite{khan2020literature}. A promising approach to alleviate these costs is to use machine learning experiments available in public online repositories, where hundreds of experiments are released to the community, with a large flow of machine learning techniques, which can be used to facilitate reuse and provide faster advancement in the area \cite{brazdil2022metalearning}.

One of the largest public repositories that enable sharing of experiments in machine learning is OpenML. It offers a wide range of datasets, experiments, and results. It makes the datasets available and provides detailed information about the experiments performed on this data, including descriptions of the pipeline blocks — such as scaling, encoding, feature preprocessing, imputation, class balancing, classification, clustering and regression algorithms, along with their corresponding hyperparameters and the evaluation metrics employed. This makes OpenML a rich metadata source, allowing researchers to explore, reproduce, and analyze results \cite{bischl2017openml}. OpenML offers a web API that makes it easy to submit new results by integrating it into popular machine learning tools \cite{van2013openml}.

Although this repository offers a wide range of datasets, the quality and representativeness of recorded experiments may vary. One problem observed is the low usage of preprocessing blocks in pipelines. Even when these blocks are applied, limited exploration of their possible techniques is observed. This happens possibly because users can perform custom preprocessing and model selection in their pipelines without logging this information to OpenML. The experiments are typically not executed on OpenML's servers and may instead be run locally \cite{brazdil2022metalearning}, which hinders the creation of complete experiment registries.
%Users can perform custom preprocessing and model selection in their pipelines, and not log this information to OpenML; experiments are typically not run on an OpenML server and can be run locally \cite{brazdil2022metalearning}. 
OpenML currently presents data about 22,298 machine learning pipelines. We carried out an analysis of these pipelines and found that only 47.09\% included at least one preprocessing block; 23.20\% of the pipelines included function transformer block, 7.70\% included the scaling block; 6.84\% included feature preprocessing; 3.93\% applied missing value imputation methods; 3.64\% employed encoding for categorical variables; only 0.16\% included data resampling techniques. Additionally, 1.60\% registered the use of preprocessing techniques but did not specify which methods were applied. This indicates that most of the records in OpenML are focused on the classifier, neglecting the preprocessing steps of the pipeline \cite{kuhn2018automatic}, \cite{perrone2018scalable}. This imbalance can induce bias in meta-learning systems that rely on OpenML for its meta-data.

Previous studies highlight the importance of preprocessing techniques \cite{obaid2019impact}, \cite{raju2020study}. A recent study investigated the impact of scaling techniques on the performance of classification algorithms, for example,  \cite{de2023choice}. Its results show the importance of this preprocessing step and how it can significantly influence model performance. While the use of preprocessing techniques is beneficial, it is essential to have a diverse dataset that covers a wide range of scenarios for preprocessing algorithms. However, building such a comprehensive and diverse collection is challenging \cite{pio2024review}.

%Ensuring representative and informative is crucial and can benefit areas such as Meta-Learning (MtL). MtL allows you to learn from previous machine learning experiences and transfer the acquired knowledge to new tasks \cite{Hutter}, mapping the characteristics (meta-features) of the datasets with information that describes the pipelines' performances.
%, and one of the area limitations it is precisely the computational cost associated with evaluating performance on datasets and meta-features extraction \cite{Khan}.

In this context, the main goal of this paper is to present
PIPES, a meta-dataset for meta-learning that we are making available to the community. PIPES aims to support meta-learning by providing researchers with a comprehensive and representative collection of results covering different pipeline blocks (classifiers and data preprocessing) evaluated on multiple datasets. To this end, several pipelines are evaluated on datasets of various sizes. By incorporating multiple preprocessing blocks, each including many possible techniques in a representative way, PIPES aims to overcome the limitations observed in OpenML and provide a more complete collection of pipelines along with their results when applied to diverse datasets,  facilitating machine learning and meta-learning research. 

This comprehensive evaluation results in a set of records that can be used to advance meta-learning research, better understand the outcomes of different combinations of machine learning models and preprocessing techniques on different datasets and allow for important insights. The main contributions of this paper are:

\begin{itemize}
     \item We present PIPES, a meta-dataset of machine learning experiments seeking completeness and diversity of the pipelines, including several preprocessing blocks and a classification block. The meta-dataset includes full details of the experiment setups and outcomes to ensure easy replication by researchers aiming to advance research in meta-learning.    
     
     \item We provide an API through which the user can fetch the meta-data and also expand PIPES. Easing interaction with the repository\footnote{\url{https://github.com/cynthiamaia/PIPES.git}}.
     
     \item We exemplify a use of PIPES in meta-learning research for pipeline recommendation.
     
     \item We analyze and compare the representativeness and completeness of pipelines from PIPES with those obtained from OpenML, one of the largest public repositories.
     
\end{itemize}

\section{Background and Related Work}
\label{related-work}
Meta-Learning enables learning from prior machine learning experiences to tackle tasks like algorithm recommendation~\cite{hutter2019automated}. This process involves two levels: base level and meta level. At the base level, algorithms and evaluation metrics are defined to evaluate their performance on datasets. At the meta level, a meta-model is trained using meta-datasets that comprise meta-examples — datasets represented by meta-features (characteristics of the dataset) and meta-targets (e.g., algorithm performances on the dataset). This trained meta-model can be used to recommend the most suitable base-level algorithm for new datasets. The type of meta-model depends on the meta-target \cite{brazdil2008metalearning}.

Meta-features, which describe datasets, fall into categories like simple, statistical, information theory, model-based, complexity-based, and performance-based (landmarking) \cite{brazdil2022metalearning}. Simple meta-features capture general dataset properties (e.g., number of classes), while statistical ones focus on measures like mean and standard deviation. Information theory meta-features assess attributes like entropy \cite{castiello2005meta}, and model-based ones reflect properties of trained models (e.g., decision tree leaf count) \cite{brazdil2008metalearning}. Landmarking meta-features evaluate the performance of simple, fast-to-train algorithms (such as Naive Bayes) on a dataset, providing an indication of the dataset's properties through the relative success of these algorithms \cite{brazdil2022metalearning}. Complexity meta-features analyze aspects like class separability and attribute overlap \cite{rivolli2022meta}. Together, these meta-features enable effective algorithm recommendations for new datasets.

In meta-learning, most of the studies have focused on recommending predictive models, such as classifiers and regressors \cite{zhu2018new, mustafa2017alors, pimentel2019new,wang2014generic, dantas2018selecting}. On the other hand, only a few studies have focused on recommending preprocessing algorithms \cite{pio2024review, khan2023autofe, avelino2024resampling, de2024meta}, and recommending hyperparameters for classifiers \cite{zhang2015multi}. Studies agree that one of the limitations of the meta-learning area is the computational cost associated with the meta-base construction phase. This cost arises because algorithms need to be executed on multiple datasets, making the process expensive \cite{pio2024review}, \cite{khan2020literature}. However, collaborative structures that allow sharing of experimental results can help reduce computational costs. Investigating how these tools are used in practice can significantly contribute to advances in the area. This type of analysis is crucial to understanding how the results obtained can effectively promote reproducibility and drive advances in meta-learning. 

%The following machine learning studies that aim at reproducibility in machine learning are discussed. 

%The Table \ref{tabela1_comparative} summarizes the techniques pre-processing employed, the classification models used, the number of datasets, and the availability of published meta-features.

Vanschoren et al. \cite{vanschoren2012experiment} present a structure aimed at facilitating the sharing of experiments in machine learning. Experiments were carried out on 84 datasets, evaluating 54 Weka algorithms. Fifty-six meta-features were calculated for each dataset, although the study does not explicitly state the types of meta-features considered. Furthermore, the experiments are not accessible, despite the link being provided in the study, they do not allow access to the structure. Nonetheless, this study was the precursor to the design of OpenML \cite{vanschoren2014openml}. OpenML is an open-source platform that allows the sharing of datasets and experiment meta-data. It is integrated into popular platforms such as Weka, R, MOA, and Scikit-learn. It offers a website with access to 5,866 datasets and 22,298 machine learning pipelines, and also provides access to some kinds of meta-features, such as simple measures, statistics, information theory, and landmarking. 

Kaggle is also an open-source platform that strives for reproducibility, but focuses on machine learning competitions and challenges \cite{bojer2021kaggle}. Unlike OpenML, which offers a robust system for recording metadata in a standardized way, Kaggle doesn't provide standardization in describing and sharing experiments, limiting its application to structured analysis and reproducibility.

NAS-Bench-101 \cite{ying2019bench} is an architecture meta-dataset designed for Neural Architecture Search (NAS). This dataset includes a comprehensive record of training and evaluation results for a wide range of Convolutional Neural Network (CNN) algorithms. The publicly available meta-dataset covers data, search space, and training code, aiming to promote reproducibility. However, unlike OpenML, which supports various machine learning tasks and pipelines, NAS-Bench-101 is focused on CNN architectures. 

While OpenML is a comprehensive repository that supports a variety of machine learning tasks, its results have limitations. One is a lack of representativeness, especially when using pipeline blocks encompassing data preprocessing techniques. Many users perform experiments locally and may not log each step of the pipeline, compromising the integrity of the model performances provided and ultimately hindering progress in the meta-learning field. Less than half (47.09\% ) of the pipelines currently available in OpenML employ at least one preprocessing block. Among these, 23.20\% uses function transformers, 7.70\% uses scaling techniques, 6.84\% uses feature preprocessing, 3.93\% applies missing value imputation and 3.64\% uses encoding strategies for categorical variables. The statistics highlight the need for greater exploration of data preprocessing techniques in experiments since many are little applied (or logged) regardless of their known importance to classification performance.

PIPES aims to provide large-scale record of machine learning experiments covering a diverse range of pipelines and datasets. It is therefore expected that it enables advances in meta-learning research, especially with regards to the effects of previously underrepresented pipeline blocks and techniques.

\section{Proposed Meta-Dataset}
\label{proposed}

A meta-dataset is an indispensable input in a meta-learning process. In the context of the algorithm selection problem, for example, a meta-learning process consists of recommending the most suitable algorithm for a specific task based on metadata from previous experiments in machine learning. In this sense, this study proposes a meta-dataset that enables the sharing of well-structured experiments, taking into account the interactions among the several pipeline blocks and on different datasets. Therefore, the proposed PIPES meta-dataset consists of a comprehensive collection of pipeline details associated with their outcomes for different datasets.

The goal is to provide the machine learning research community with a valuable meta-dataset, which can be used to train models in a holistic way, to recommend complete pipelines (i.e., including the predictive model and the data preprocessing blocks). Since the models will be trained on a large meta-dataset that is built with a focus on completeness, it is expected that they will present a high generalization power. This way, PIPES can help the community identify the most promising pipelines to solve specific problems based on frequency of use and performance in previous experiments. This knowledge can help in designing more effective and targeted search spaces.

Formally, the construction of the PIPES meta-dataset is defined as follows. Consider the set of datasets used,
$\mathbb{D} =  \{\mathbf{D}_1, \mathbf{D}_2, \ldots, \mathbf{D}_z\}$. For each dataset \(\mathbf{D}_i \in \mathbb{D} \), all the different pipelines are applied, with each pipeline consisting of a chain of blocks (steps) that each have several possible techniques. 

Given the following set definitions:
\vspace{-0.1cm}
\begin{align*}
\mathbb{T} &= \{ t_1, t_2, \dots, t_k \} \text{: The set of imputation techniques. } \\
\mathbb{E} &= \{ e_1, e_2, \dots, e_l \} \text{: The set of encoding techniques.} \\
\mathbb{P} &= \{ p_1, p_2, \dots, p_m \}\text{: The set of scaling techniques.} \\
\mathbb{A} &= \{ a_1, a_2, \dots, a_n \} \, \text{: The set of feature preprocessing,} \\
&\text{which involves transformation and feature selection.} \\
\mathbb{C} &= \{ c_1, c_2, \dots, c_p \}\text{: The set of classifiers.}
\end{align*}

\begin{figure}[!ht]
\centering
%recortar figura (remover bordas): trim=left bottom right top, clip  
\includegraphics[width=0.3\textwidth]{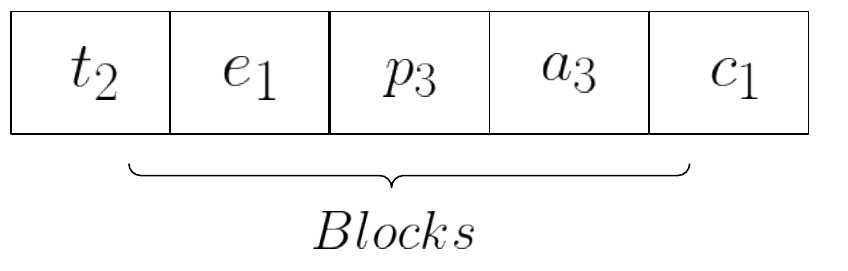} 
\vspace{-0.2cm}
\caption{Representation of an example of a specific pipeline.}
\label{fig:pipeline}
\end{figure}

\noindent A pipeline is a tuple containing five elements (blocks). The first block is populated by an imputation technique $t \in \mathbb{T}$, the second block corresponds to a categorical encoding technique $e \in \mathbb{E}$, then comes the pipeline block responsible for data scaling with a technique $p \in \mathbb{P}$, followed by a feature preprocessing techinique $a \in \mathbb{A}$, and, finally a classification algorithm $c \in \mathbb{C}$. In Fig. \ref{fig:pipeline}, we illustrate an example of a specific pipeline.

\vspace{-0.3cm}

%The justification for selecting these sets of techniques is defined in Section \ref{experiments}, while Table \ref{tab:classification-preprocessing}, also in Section \ref{experiments} , provides the complete list of all techniques. 
Thus, for each dataset \( \mathbf{D}_i \in \mathbb{D} \), the set of possible pipelines \( \mathbb{S} \) is given by the combination of all elements of the sets \( \mathbb{T}, \mathbb{E}, \mathbb{P}, \mathbb{A} \) and \( \mathbb{C} \), that is:

\begin{equation}
\label{eq:S}
\mathbb{S} = \mathbb{T} \times \mathbb{E} \times \mathbb{P} \times \mathbb{A} \times \mathbb{C}
\end{equation}

Finally, considering that each dataset \( \mathbf{D}_i \) is represented by a vector \( \mathbf{f}_i = \{f_i^{(1)}, f_i^{(2)} , \ldots, f_i^{(x)} \} \), with \( x \) meta-features that characterize its properties, the meta-dataset $\mathbb{S}^*$ is generated, bringing together all pipelines and the results of each dataset in each fold, covering all possibilities, not limited to just the best performance. Thus, we have the following formulation:

\footnotesize
\begin{equation}
\mathbb{S}^* = \left\{ 
\left( \mathbf{f}_i, S_j, H, {\tau_{\text{train}}}, \tau_{\text{test}}, \epsilon \right)
\mid
\begin{array}{l}
\forall i \in \{1,2,3,\dots,z\}, \\
\forall S_j \in \mathbb{S},\\
H, \tau_{\text{train}}, \tau_{\text{test}} \in \mathbb{R}
\end{array}
\right\}
\label{eq:meta-dataset}
\end{equation}

Where $\mathbb{R}$ is the set of real numbers. Therefore, the set $\mathbb{S}^*$ contains a detailed collection of experiments using different pipelines, including dataset characteristics $\mathbf{f}_i$, the achieved performance $H$, the training and testing times $\tau_{\text{train}}$ and $\tau_{\text{test}}$ and a text field with information about errors ($\epsilon$) occurred during processing, if any. These errors may refer, for example, to situations where a dataset contains missing values and, in certain pipelines, the imputation technique is absent. 

%By exploring these experiments, it is possible to verify how execution times and errors occur in certain datasets and pipelines. $H$ corresponds to all pipeline predictions that were collected, enabling the calculation of any performance metric.

%\subsection{API design}

%One of the main objectives of MetaD-MLP is to contribute to the community by providing reproducible results in the area and promoting the advancement of research. In Figure \ref{fig:metadmlp}, we illustrate how users can retrieve and insert more metadata. The projected initial use of MetaD-MLP\footnote{\url{https://anonymous.4open.science/r/MetaD-MLP-5DFB}} will be through an API (Application Programming Interface). The Datasets component contains the entire collection of metadata stored in the backend, while data retrieval and insertion requests are handled through the API.

%\begin{figure*}[!ht]
%\centering
%\includegraphics[width=1.0\textwidth]{design_metad.drawio (8).pdf}
%\vspace{-0.2cm}
%\caption{API design of MetaD-MLP.}
%\label{fig:metadmlp}
%\vspace{-0.4cm}
%\end{figure*}

%An example demonstrates this interaction: A POST request sends new metadata, including the dataset ID, name, and location. Afterward, a GET request retrieves all metadata, such as dataset ID, fold information, training and testing times, and a specific metric with its corresponding value. This API-driven approach makes it easy to interact with the metadata repository. The front-end interface will be designed for better user interaction with metadata, and everything will be connected through GitHub.

\section{Meta-dataset Construction}
\label{experiments}
In this section, we detail the procedure performed to build the proposed meta-dataset according to the formal description in Section \ref{proposed}.

\subsection{Datasets}
\label{datasets}

The datasets were collected from the open-source OpenML platform in September 2024. We selected only datasets with the number of instances in the range [100, 100000], the number of classes in [2, 100], and the number of attributes in [5, 100]. Initially, the procedure resulted in a total of 1,248 datasets. From these datasets, 867 were excluded to reduce the number of datasets belonging to the same family and to remove repeated datasets the had been logged with different names, resulting in 381 distinct datasets.
%, with greater diversity in different domains. 
From the 381 datasets selected, 273 are binary-class problems, and 108 are multi-class problems. The final list of datasets used is presented in the supplementary material. %In this initial study, out of 381 datasets, 221 were used, while the rest are still being executed.

%\subsection{Evaluation Metric}

%The datasets were evaluated using 5-fold cross-validation and accuracy was used as the initial evaluation metric. However, all combination predictions were collected; other metrics can also be evaluated, such as precision and f-score.

\subsection{Pipeline blocks}
\label{searchspace}
Given the wide variety of different classification algorithms and preprocessing methods, we limited the selection of the techniques to populate the pipeline blocks based on the methods used in Auto-Sklearn \cite{feurer2015efficient}, which uses the Scikit-learn package \cite{pedregosa2011scikit}. The pipelines' blocks and their possible techniques are presented in Table \ref{tab:classification-preprocessing}. Note that each preprocessing block includes an option of not being executed (None), while the classification block is the only one that is mandatory.

%\vspace{-0.3cm}
\begin{table}[ht]
\centering
\caption{Pipeline blocks and their possible techniques.}
\label{tab:classification-preprocessing}

%\vspace{2ex} 

\hspace{0.1cm}
\vspace{2.0ex} 
\begin{minipage}{0.2\textwidth}
\centering
Imputation ($\mathbb{T}$) \\[1ex]
\resizebox{0.8\textwidth}{!}{%
\begin{tabular}{l}
\hline
%Name \\ \hline
SimpleImputer(SI), None.\\ \hline
\end{tabular}%
}
\end{minipage}
\hspace{0.1cm}
\vspace{2.0ex} 
%\hspace{0.1\textwidth} % Ajuste o espaço entre as tabelas
% Tabela E (direita)
\begin{minipage}{0.42\textwidth}
\centering
Encoding ($\mathbb{E}$) \\[1ex]
\resizebox{0.75\textwidth}{!}{%
\begin{tabular}{l}
\hline
%Name \\ \hline
OrdinalEncoder (OE), OneHotEncoder (OHE), None.\\ \hline
\end{tabular}%
}
\end{minipage}
\hspace{0.4cm}
\vspace{2.2ex} 
\begin{minipage}{0.42\textwidth}
\centering
Scaling ($\mathbb{P}$) \\[1ex]
\resizebox{0.85\textwidth}{!}{%
\begin{tabular}{l}
\hline
%Name \\ \hline
MinMaxScaler (MM), StandardScaler (SS), PowerTransformer (PT), \\ QuantileTransformer (QT), RobustScaler (RS), Normalizer (Nor), None.\\ \hline

\end{tabular}%
}
\vspace{1ex} 
\end{minipage}
\hspace{0.1cm}
\vspace{0.3ex} 
% Tabela A
\begin{minipage}{0.42\textwidth}
\centering
Feature Preprocessing ($\mathbb{A}$) \\[1ex]
\resizebox{0.9\textwidth}{!}{%
\begin{tabular}{l}
\hline
%Name \\ \hline
ExtraTrees prep (ETP), FastICA (FICA), Nystroem (NY), \\
FeatureAgglomeration (FAGG), GenericUnivariateSelect (GU),\\
LinearSVC prep (LSVCP), Principal component analysis (PCA), \\
KernelPCA (KPCA), Radial Basis Function Sampler (RBFS), \\
PolynomialFeatures (PF),RandomTreesEmbedding (RTE),\\
SelectPercentile (SP), TruncatedSVD (TSVD), None.\\ 
\hline
\end{tabular}%
}
\end{minipage}

\vspace{2ex} 

\begin{minipage}{0.41\textwidth}
\centering
Classification ($\mathbb{C}$) \\[1ex]
\resizebox{1\textwidth}{!}{%
\begin{tabular}{l}
\hline
%Name \\ \hline
AdaBoost (AB), BernoulliNB (BNB), DecisionTree (DT),\\ 
ExtraTrees (ET), GaussianNB (GNB), HistGradientBoosting (HGB), \\   K-Nearest Neighbors (KNN), LinearDiscriminantAnalysis (LDA), \\ LinearSVC (LSVC), Multi-layer Perceptron (MLP),  \\
MultinomialNB (MNB),  QuadraticDiscriminantAnalysis (QDA), \\
PassiveAggressive (PA),  Support Vector Classification (SVC), \\
Stochastic Gradient Descent (SGD), RandomForest (RF). \\
\hline
\end{tabular}%
}
\end{minipage}

\vspace{-0.1cm} 
\end{table}

This results in a total of 9,408 combinations, i.e., pipelines, (2 imputation techniques $\times$ 3 categorical encoders $\times$  7 scalers $\times$ 14 feature preprocessing techniques $\times$ 16 classification algorithms). We consider the hyperparameters in their default values for this first analysis; however, we highlight the potential to extend our meta-dataset in future work to also address hyperparameter variations. Apart from these techniques, Auto-sklearn also incorporates a class balancing method that is embedded within AutoML system and not as easy to replicate. Therefore, we focus on techniques and algorithms that are easily accessible and configurable in Scikit-learn. %The constructed blocks and training and testing time are recorded for each pipeline tested in the dataset. This allows us to evaluate the performance of each pipeline specifically.

Following the definition in Eq. \ref{eq:S}, all pipeline possibilities are explored. Some of these pipelines may be invalid or lead to errors for some datasets. Nonetheless, all of them were documented and included in the meta-dataset. The meta-dataset, therefore, covers all results, not just the best one for each dataset, allowing analyses of the performance of all pipelines across all datasets.

\subsection{Meta-Features}

The extracted meta-features belong to six groups: simple (12 features), information theory (13 features), statistics (48 features), model-based (24 features), landmarking (14 features), and complexity (34 features), with a total of 145 meta-features, which are detailed in the supplementary material.

\subsection{Hardware and Software} 

We employed three computing clusters and ran all the pipelines on 300 datasets through several parallelized jobs, each job running a fraction of the pipelines on a given dataset. After the executions were completed, the results were combined into a structured table containing the various details necessary for composing the meta-dataset. We used the Python language (version 3.9.1) and the following packages: Scikit-Learn (version 1.2.2) \cite{pedregosa2011scikit}, (version 1.5.3), OpenML (version 0.15.0) \cite{feurer2021openml}, and PyMFE (version 0.4.2) \cite{alcobacca2020mfe} .

%from the Digital Research Alliance of Canada\footnote{\url{https://docs.alliancecan.ca/wiki/Technical_documentation}}: Narval, Beluga and Graham. We ran all the pipelines on 221 datasets through several parallelized jobs, each job running a fraction of the pipelines on a given dataset. 

\section{Analysis}

%colcoar illustration de usod e meta-learning ...
This section details the experiment performed and seeks to answer the following research questions:

\begin{description}
    \item[RQ1] Does PIPES overcome OpenML's limitations regarding the biases and data imbalance of pipelines and contribute to the advancement of meta-learning? 
    \item[RQ2] Are the datasets selected in PIPES diverse?% that contribute to advancements in meta-learning
\end{description}

To answer RQ1, we perform two analyses that we present in subsections \ref{sectionA} and \ref{sectionB}. First, we perform an exploratory analysis of the pipelines available in OpenML (subsection \ref{sectionA}). The idea is to study how frequently each technique is used within each pipeline block. Then, in subsection \ref{sectionB}, we present an example of using PIPES for a meta-learning task to recommend the optimal techniques for Feature Preprocessing and Scaling, given a fixed classifier, the SVC.  These two particular blocks and the classifier were chosen because of their high frequency in OpenML data, based on the outcomes of the exploratory analysis in subsection \ref{sectionA}. We compare the classification performance achieved using the optimal pipeline from OpenML data versus PIPES data. Finally, in subsection \ref{sectionC} we answer RQ2 by studying the diversity of the PIPE's datasets according to their meta-feature representations.

\subsection{Exploratory analysis of the pipelines from OpenML}
\label{sectionA}

\begin{figure*}[!ht]
\centering
\includegraphics[width=1.0\textwidth]{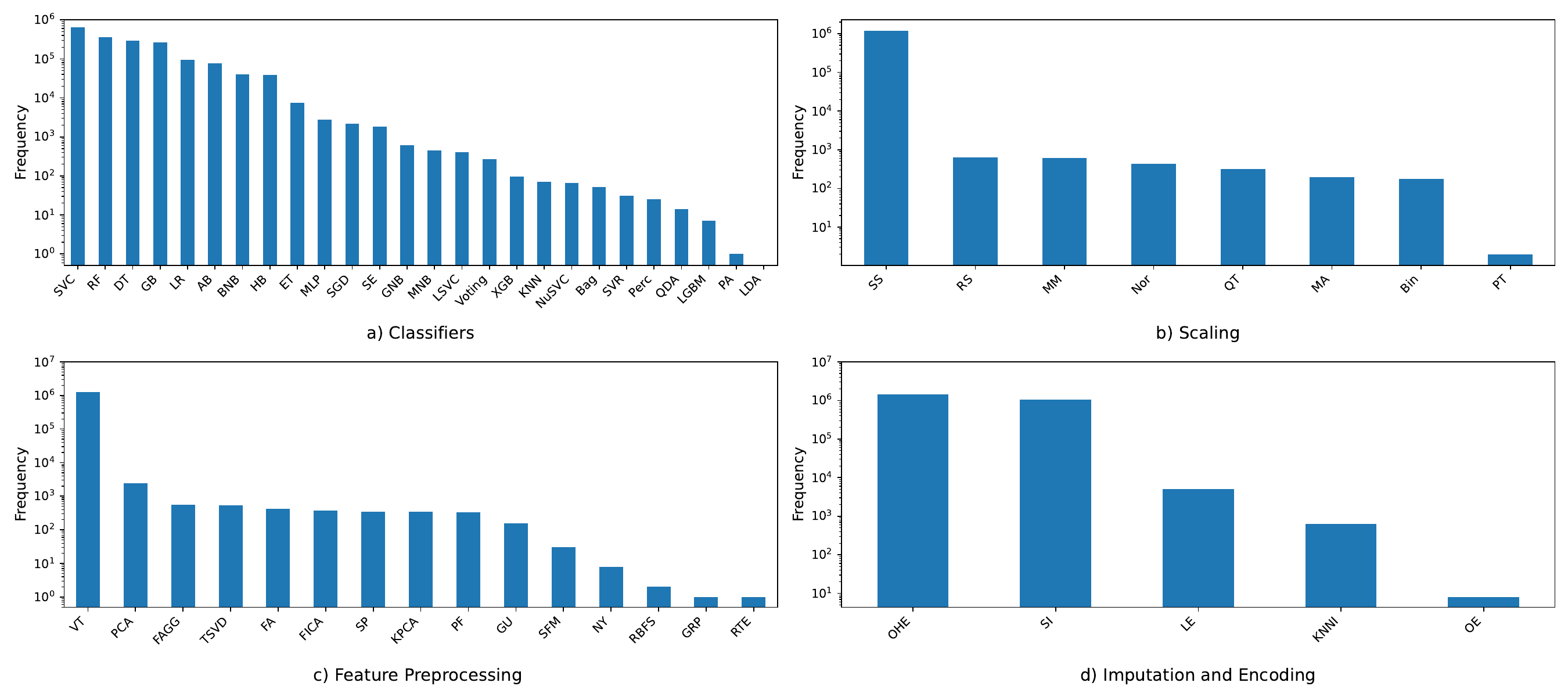}
\vspace{-0.8cm}
\caption{Frequency of execution of each technique, per pipeline block, according to  OpenML records. Most of the abbreviations for FP and scaling are given in Table \ref{tab:classification-preprocessing}. Additional abbreviations: Classifiers:  LR - LogisticRegression, XGB - XGBoost,  NuSVC - Nu-Support Vector Classification, LGBM - LightGBM,  Bag - Bagging, Perc - Perceptron, SE - StackingEstimator, Voting - VotingClassifier. Scaling: Bin - Binarizer.}
\label{fig:runs_}
\vspace{-0.4cm}
\end{figure*}

In this first analysis, we explore the pipelines recorded in OpenML w.r.t. the frequencies of each block's techniques. In Fig. \ref{fig:runs_}, the bar plots show these frequencies for all five pipeline blocks covered in our study. These graphs consider 6,729,117 classification experiments using Scikit-learn present in OpenML records.

In Fig. \ref{fig:runs_}(a), we observe that a large number of classifiers have been logged to OpenML, but their frequencies show that while some models have been used more than 100,000 times (SVC, RF, DT, GB), the majority of the models appear for less than 1,000 times, with some extreme cases presenting less than 100 examples. There is clearly a preference for SVM and decision-tree-based algorithms. The same kind of imbalance can be observed for the remaining blocks. We highlight the scaling block, where StandardScaler is employed approximately 1 million times, while the remaining techniques appear less than 1,000 times, with PowerTransformer being used less than 10 times. Moreover, upon further inspection of the data, we note that 87\% of the examples using MinMaxScaler use only six datasets, 83\% of RobustScaler's appearances occur using only seven datasets, and 100\% of PowerTransformer's examples use just one single dataset. This concentration means that, although there are many experiments available in OpenML's records, there is a limitation in the diversity of contexts in which these techniques are applied. This imbalance in the representativeness of machine learning pipeline blocks can induce bias and negatively affect the quality of meta-learning models that rely on this repository for meta-data.

\begin{figure}[!ht]
\centering
\includegraphics[width=0.48\textwidth]{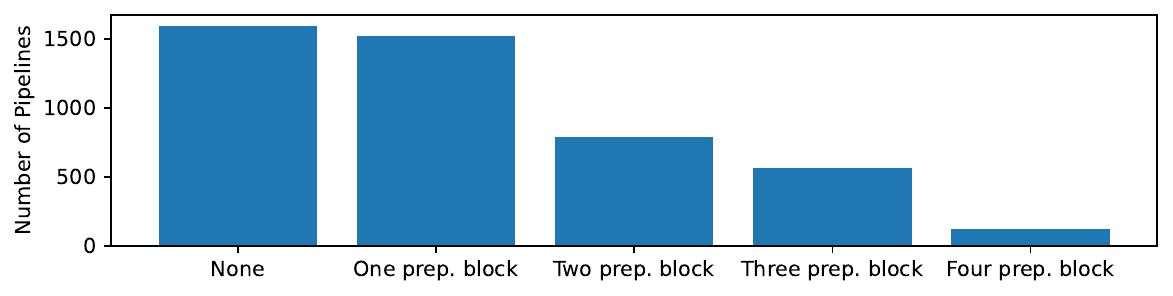}
\vspace{-0.3cm}
\caption{Quantity of pipelines that employ each number of preprocessing blocks. Considering all OpenML classification pipelines that use Scikit-learn.}
\label{fig:num_prep_blocks}
\vspace{-0.3cm}
\end{figure}

Another important aspect is the lack of diversity in the combinations of preprocessing techniques in the pipelines. Fig.~\ref{fig:num_prep_blocks} shows that most of the classification pipelines recorded on OpenML use only one or no preprocessing block at all. As the number of preprocessing blocks used increases, the number of pipelines decreases. Only 2.74\% of the pipelines employ all four pipeline blocks. This leads to bias in the way the blocks are combined. For example, VarianceThreshold, which is the most frequent Feature Preprocessing technique is, in most cases, associated with the StandardScaler scaling technique, while FeatureAgglomeration is only used with StandardScaler in two examples, and KernelPCA, Nystroem, FastICA and TruncatedSVD are never combined with scaling techniques. %And most combinations were combined with OneHotEncoder. 
These observations reflect a limited exploration of different combinations of techniques within pipelines and the use of preprocessing techniques in OpenML records. 

Therefore, the main limitations observed in OpenML records include a large concentration of examples using a few of the techniques within each block, the lack of exploration of the possible combinations of pipeline blocks, and a large portion of the examples using just a few datasets. PIPES addresses these limitations by exploring pipeline blocks in depth, with greater representation and diversity in use. In our proposed PIPES meta-dataset, every possible pipeline is applied to every dataset. All the techniques within each pipeline block are equally explored, providing a balanced meta-dataset, which allows for an unbiased training of recommendation meta-models.

\subsection{Comparing PIPES and OpenML in a meta-learning task}
\label{sectionB}

In this subsection, we present an example comparing the use of PIPES and OpenML in a meta-learning task, contributing to answering RQ1. The meta-learning task in this example is to automatically recommend the optimal feature preprocessing and scaling techniques for the SVC classifier, given a particular dataset represented by its meta-features vector. In this example, two meta-datasets are built: one based on PIPES' results and the other using OpenML results. Subsequently, these meta-data sets are compared to assess their representativeness. The goal is to examine the registered pipelines and their representativeness in using feature preprocessing and scaling blocks. These two preprocessing blocks and the SVC classifier were selected based on the analysis outcomes in subsection~\ref{sectionA}, choosing the most commonly used blocks in OpenML to retain a substantial amount of learning data, maintaining a fair comparison. We also restrict the datasets to only the 192 that appear in both PIPES and OpenML.

The OpenML meta-dataset, $\mathbf{M_{\text{OpenML}}}$, used for this analysis is composed as follows. First, $\mathbb{D}_{\text{selected}}$ contains all 192 datasets in common with PIPES. After, given the set $\mathbb{F}$ of pipelines that use SVC as the classifier, defined as $\mathbb{F} = \{ \mathbb{\sigma}_1, \mathbb{\sigma}_2, \ldots, \mathbb{\sigma}_n \}$, we take only the pipelines that are the best for each selected dataset to form $\mathbb{F}_{\text{best}}$, defined in Eq. \ref{eq:F_best}, which constitutes our meta-target.

\begin{equation}
\label{eq:F_best}
\mathbb{F}_{\text{best}} = \left\{ \mathbb{\sigma}_{\text{best}}(\mathbf{D}_i) | \forall \mathbf{D}_i \in \mathbb{D}_{\text{selected}} \right\}
\end{equation}

Then, the meta-features vectors \( \mathbf{f}_i \) representing each dataset \( \mathbf{D}_i \) are merged with the meta-targets. The resulting meta-dataset $\mathbf{M}_{\text{OpenML}}$ is composed of the tuples $(\mathbf{f}_i, \mathbb{\sigma}_{\text{best}}(\mathbf{D}_i))$, allowing the recommendation of pipeline based on the datasets' meta-features, as shown in Eq.~\ref{eq:MopenML}.

\footnotesize
\begin{equation}
\label{eq:MopenML}
\mathbf{M_{\text{OpenML}}} = \{ (\mathbf{f}_1, \mathbb{\sigma}_{\text{best}}(\mathbf{D}_1)), (\mathbf{f}_2, \mathbb{\sigma}_{\text{best}}(\mathbf{D}_2)), \dots, (\mathbf{f}_z, \mathbb{\sigma}_{\text{best}}(\mathbf{D}_z)) \}
\end{equation}

We define $\mathbf{M}_{\text{PIPES}}$ following the same procedure described for OpenML but using PIPES' best pipelines for each dataset.

\begin{equation}
\footnotesize
\mathbf{M_{\text{PIPES}}} = \{ (\mathbf{f}_1, \pi_{\text{best}}({\mathbf{D}_1})), (\mathbf{f}_2, \pi_{\text{best}}({\mathbf{ D}_2})), \dots, (\mathbf{f}_z, \pi_{\text{best}}({\mathbf{D}_z}))\}
\end{equation}

\noindent Where each pair $(\mathbf{f}_i, \pi_{\text{best}}({\mathbf{D}_i})$ represents the meta-features of the dataset $\mathbf{D}_i$ and the pipeline $\pi_{\mathbf{D}_i}$  which provided the best performance for dataset $\mathbf{D}_i$.

\begin{figure*}[!ht]
\centering
\includegraphics[width=1.0\textwidth]{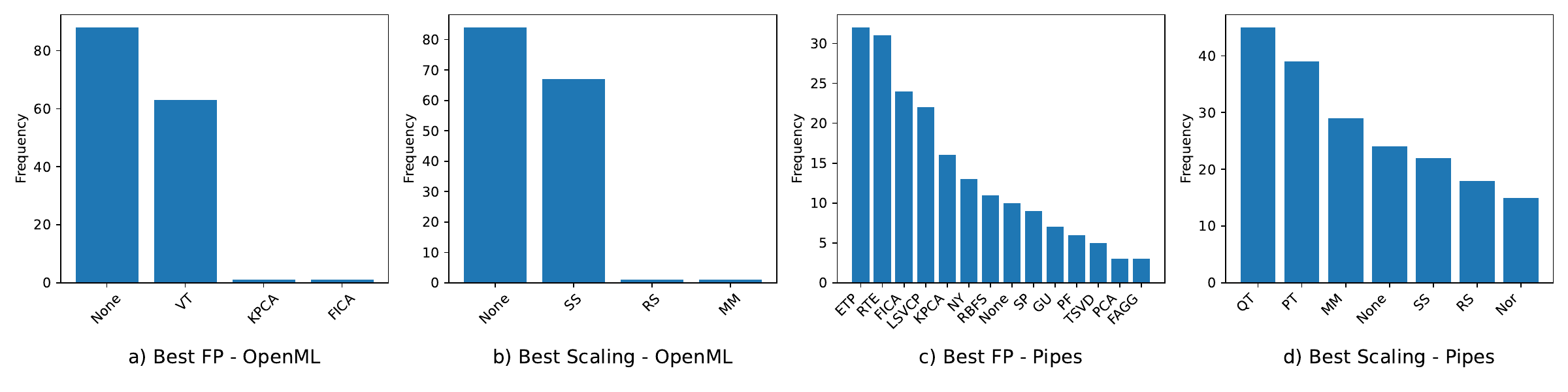}
\vspace{-0.6cm}
\caption{The graphs show how frequently each technique appears in the best pipelines for the 192 datasets in common. Most abbreviations for FP and scaling are given in Table \ref{tab:classification-preprocessing}. Additional abbreviations: VT - VarianceThreshold, FA - FactorAnalysis, SFM - SelectFromModel. Scaling: MA — Max Absolute Scaler.}
\label{fig:openml_metadmlp}
\vspace{-0.4cm}
\end{figure*}

In Fig. \ref{fig:openml_metadmlp}, the frequency with which each FP and Scaling technique appear in the pipelines from the meta-datasets is presented. Fig. \ref{fig:openml_metadmlp}a and Fig. \ref{fig:openml_metadmlp}c show the results for $\mathbf{M_{\text{OpenML}}}$ and Fig. \ref{fig:openml_metadmlp}d and Fig. \ref{fig:openml_metadmlp}c for $\mathbf{M_{\text{PIPES}}}$. Notice how the best technique indicated using OpenML data is much less varied and distributed than indicated by PIPES results. The careful analysis of the results from meta-dataset  $\mathbf{M_{\text{PIPES}}}$ can also reinforce the use of certain techniques over others according to the selected algorithm, which can help create curated search spaces for optimization processes, reducing computational cost and increasing performance for recommending pipelines.

One possible explanation for the pipeline imbalance obtained from OpenML is incomplete pipeline reporting. Many users run experiments locally, and many may not report each step of the pipeline, harming the reliability of the records.

Considering the accuracy obtained by each pipeline on the 192 datasets using 5-fold cross-validation, we compare the two meta-datasets in Table \ref{tab:comp_PIPES_OpenML}, where we also show the number of wins, ties, and losses of PIPES versus OpenML, the mean ranking, and the $p$-value from a Wilcoxon signed-rank test comparing these rankings. Notice how the PIPES pipelines obtain a higher accuracy of 0.84 versus 0.73 from the OpenML pipelines. Additionally, PIPES only lost 3 times against OpenML, winning in 93\% of the datasets. Consequently, the mean ranking of PIPES is 1.04, against 1.96 for OpenML (lower is better). The $p$-value obtained refutes the hypothesis that $\mathbf{M}_{\text{PIPES}}$ and $\mathbf{M}_{\text{OpenML}}$ have similar average rankings.

\begin{table}[!ht]
\centering
\caption{A analysis of the mean accuracy, Wins/ties/losses of the $\mathbf{M}_{\text{PIPES}}$ against $\mathbf{M}_{\text{OpenML}}$, mean ranking, $p$-value of the Wilcoxon signed rank test ($\alpha =$ 0.05.)}
\begin{tabular}{lcc}
\hline
             & \multicolumn{1}{l}{$\mathbf{M}_{\text{PIPES}}$} & \multicolumn{1}{l}{$\mathbf{M}_{\text{OpenML}}$} \\ \hline
Mean acc.    & \textbf{0.840}                                 & 0.735                                           \\ \hline
Win/tie/loss & \multicolumn{2}{c}{180/9/3}                                                                        \\ \hline
Mean rank    & \textbf{1.039}                                  & 1.960                                           \\ \hline
$p$-value    & \multicolumn{2}{c}{1.021e-31}                                                                      \\ \hline
\label{tab:comp_PIPES_OpenML}
\end{tabular}
\end{table}

Now, answering \textbf{RQ1: Does PIPES overcome OpenML's limitations regarding the biases and data imbalance of pipelines and contribute to the advancement of meta-learning?} Yes, PIPES' strategy of completeness-oriented meta-dataset construction, which gives each technique the same opportunity in the records, allows for less biased and more balanced meta-model learning that can potentially lead to new insights, contributing to the advancement of meta-learning.

\subsection{PIPES' datasets diversity}
\label{sectionC}

\begin{figure}[!ht]
\centering
%recortar figura (remover bordas): trim=left bottom right top, clip  
\includegraphics[width=0.45\textwidth]{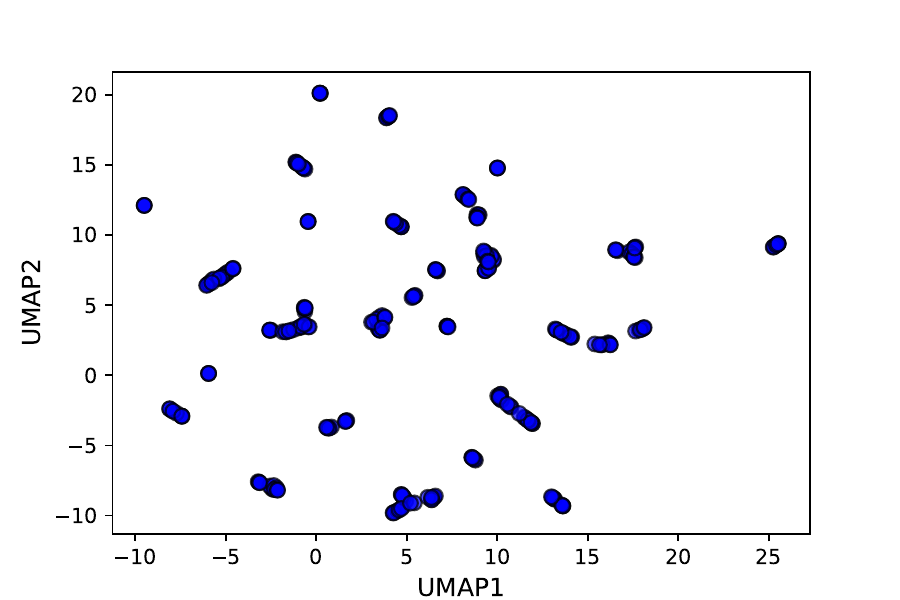} 
\caption{PIPES' datasets organized according to a UMAP representation of their meta-features' space.}
\label{fig:datasets}
\vspace{-0.3cm}
\end{figure}

To answer \textbf{RQ2: Are the datasets selected in PIPES diverse?}, in Fig. \ref{fig:datasets}, the 280 datasets from PIPES are represented in terms of all their meta-features after a transformation to a bi-dimensional space using Uniform Manifold Approximation and Projection for Dimension Reduction (UMAP)\cite{mcinnes2018umap}, configured with $k$ = 3 neighbors. Due to computational failures during the execution process, only 280 datasets could be successfully processed and included in the analysis. We observe that the datasets tend to cover the extent of the space rather than appearing in a cluster, which indicates diversity.  A meta-dataset composed of diverse pipelines contributes to the meta-learning area, as it allows the exploration of different combinations of pipeline blocks, taking into account the specific characteristics of each dataset, representing different possible real-life scenarios.

%\vspace{-0.24cm}

\section{Limitations}

PIPES offers a broad meta-dataset but is not without limitations. The main drawbacks are: (i) it does not take hyperparameters into account, (ii) it misses some other common pipeline blocks, such as data balancing. Additionally, its current architecture is not yet completely finalized, and improvements are needed, mainly in integration, to facilitate user visualization and metadata access.

\section{Conclusion}

We proposed PIPES, a meta-dataset for meta-learning. PIPES is designed for the machine learning community to enable meta-learning experimentation on a rich collection of experiments that does not neglect important pipeline blocks, particularly preprocessing, such as scaling, feature preprocessing, imputation, and encoding. Given the difficulty of obtaining collections of representative experiments from OpenML, one of the most significant learning and machine repositories, the idea is that PIPES can contribute to the reproducibility of experiments on a representative meta-dataset covering a wide range of techniques. 

PIPES can be used in the meta-learning area for tasks such as recommending the most suitable pipeline for specific datasets, analyzing the impact of different block combinations across various scenarios, and identifying promising pipelines to refine search spaces in recommendation systems. Its main advantages include broad coverage of problem domains, representative pipeline blocks (covering classifiers and preprocessing techniques), and detailed training/testing time data for cost-benefit analysis. In addition, it provides all predictions, allowing researchers to apply various evaluation metrics. For future work, we intend to optimize the algorithms' hyperparameters, complete the execution of the datasets, add more data and pipeline blocks, and also create an accessible tool to facilitate users' insertion and retrieval of data.

\end{document}